\documentclass[runningheads]{llncs}

\usepackage[T1]{fontenc}

\usepackage{algorithm}
\usepackage{algpseudocode}
\usepackage{amsmath}
\usepackage{amssymb}
\usepackage{booktabs}
\usepackage{color}
\usepackage{graphicx}
\usepackage{hyperref}
\usepackage{letltxmacro}
\usepackage{subfig}
\usepackage{tabularray}

\UseTblrLibrary{booktabs}

\pdfpagesattr{/CropBox [86 65 526 731]}

\gdef\papertitle{Revisiting Structural Dependency in Autoregressive Multi-Task Table Recognition via Order-Independent Cell-Level Representations}
\title{\papertitle}
\titlerunning{Structural Dependency in Autoregressive Multi-Task Table Recognition}

\author{Takaya Kawakatsu\inst{1}}
\authorrunning{T. Kawakatsu}

\institute{Preferred Networks, Inc., 1-6-1 Otemachi, Chiyoda, Tokyo, Japan.\\
\email{kat.nii.ac.jp@gmail.com}\\
\url{https://researchmap.jp/t.kat}
}

\hypersetup{pdftitle={\papertitle},pdfauthor={Takaya Kawakatsu},hidelinks}

\LetLtxMacro\oldeqref\eqref
\renewcommand{\eqref}[1]{Eq.~\oldeqref{eq:#1}}

\NewDocumentCommand\algoref{m}{Algorithm~\ref{alg:#1}}
\NewDocumentCommand\secref{m}{Section~\ref{sec:#1}}
\NewDocumentCommand\figref{m}{Fig.~\ref{fig:#1}}
\NewDocumentCommand\tabref{m}{Table~\ref{table:#1}}

\NewDocumentCommand\CM{}{\checkmark}
\NewDocumentCommand\TED{}{\mathrm{TED}}

\NewDocumentCommand\TD{}{\texttt{<td>}}

\begin{document}
\maketitle

\begin{abstract}
Multi-task table recognition jointly addresses table structure prediction, cell localization, and cell content recognition within a unified framework.
Existing approaches often rely on autoregressive decoders to generate table structures and reuse their hidden states for cell localization and content recognition.
This autoregressive generation process can make cell representations order-dependent, degrading global consistency across cells.
This paper proposes a structural refinement module that produces order-independent cell features through non-causal attention.
This design enables parallel inference of cell contents while conditioning each cell on global context encoded in the refined features.
Experiments on two large datasets demonstrate consistent gains in cell localization and end-to-end recognition, while reducing overall inference time by around threefold.
\end{abstract}

\section{Introduction}

Table recognition aims to convert table images into structured formats, such as HTML, which preserve both geometric layout and textual cell contents.
Modern methods~\cite{Kat24,Nam23LA,Nam23GA,Nam23WS} use multi-task models to jointly predict table structure, cell locations, and cell contents.
Sharing structural features provides helpful context for cell localization and content recognition.

In practice, many methods~\cite{Kat24,Nassar22,VCG21,Zhong20} rely on autoregressive (AR) decoders to generate structural sequences.
However, the causal attention mask allows each cell representation to access only previously generated cells, leaving information from later cells unavailable.
The resulting representations reflect generation order rather than intrinsic table structure.
The reuse of such order-dependent features in downstream tasks may weaken the global consistency of cell relationships.

In this paper, we revisit multi-task table recognition from the perspective of structural representation design.
Instead of modifying tokenization strategies or decoder architectures, we separate structural generation from cell-level structural aggregation.
We introduce a global structural refinement module which removes generation-order dependency at the cell level.
Cell representations extracted from the structural decoder are refined through non-causal self-attention and realigned with visual features, producing order-independent structural features.

\begin{figure*}[tb]
\centering
\includegraphics[width=\textwidth]{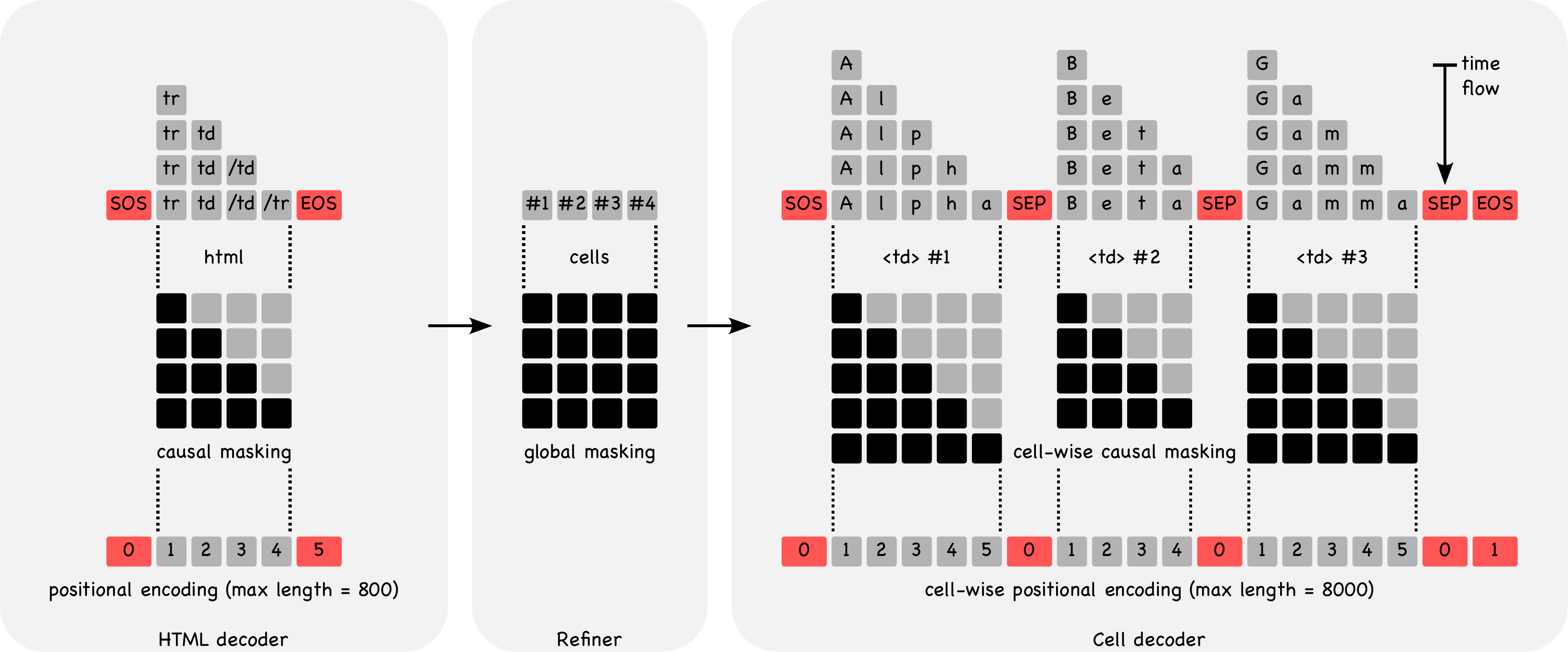}
\caption{Key idea of order-independent cell features and parallel cell decoding.}
\label{fig:idea}
\end{figure*}

As illustrated in \figref{idea}, causal masking appears in both structural decoding and cell-content decoding.
In these stages, the triangular mask ensures that each token is conditioned only on its past decoding context.
The proposed refinement module bridges these two processes by applying non-causal attention to the cell sequence.
It transforms the order-dependent cell features into globally aggregated cell representations.
These refined representations provide the basis for improved geometric consistency and parallel cell-content inference.

Experimental results on two large-scale table datasets~\cite{Zheng21,Zhong20} show consistent improvements over prior multi-task approaches.
The proposed method improves cell localization intersection over union (IoU) by around 3--5 points and increases tree edit distance-based similarity (TEDS), particularly for tables with long cell contents.
Furthermore, aggregating structural context before content generation enables parallel cell-level decoding, which achieves a roughly threefold inference speedup over MuTabNet~\cite{Kat24} while maintaining comparable recognition accuracy.

These results demonstrate the importance of dependency modeling in shared structural representations and suggest that dependency topology is a promising design axis for structured document recognition.

\section{Related Work\label{sec:related}}

Table recognition predicts structural elements such as rows, columns, cells, and span attributes from a table image, localizes each cell region, recognizes the text inside each cell, and outputs a table representation.
Prior studies can be viewed from three perspectives: geometric detection, image-to-sequence generation, and multi-task modeling.

Geometric methods~\cite{Ito93,Kien98,Liu21,Pra20,Qiao21,Raja20,Raja22,Sch17,Wang04,Zhang22} detect table components such as rows, columns, text blocks, or cells, and infer the table structure from spatial relationships.
These methods are effective when table and component boundaries are visually clear, but can be sensitive to ambiguous cell boundaries and complex merged cells.

Image-to-sequence models~\cite{Deng17,Deng19,Zhong20} generate LaTeX or HTML sequences from table images using encoder--decoder architectures.
Recent studies~\cite{Nassar22,VCG21,Huang23} have extended this formulation with Transformer models.
This top-down formulation avoids reliance on hand-designed reconstruction rules, but it linearizes the table structure into a one-dimensional sequence, making structural features dependent on the chosen generation order.

Multi-task frameworks~\cite{Zhong20,Nam23GA,Nam23WS,Nam23LA,Kat24} jointly model structure prediction, cell localization, and cell content recognition.
These frameworks use shared structural features to promote consistency among logical structure, visual layout, and text recognition.
Nevertheless, the shared structural features are typically reused from AR decoder states, leaving the causal dependency among cells unresolved.

Modern vision--language models (VLMs)~\cite{Kim22,Lee22,Wan24,Zhang24} provide strong reasoning capabilities for document understanding.
However, many of them rely on broader training and task settings that differ from standard evaluation protocols for table recognition.
Our focus is on end-to-end table recognition without using external OCR or additional training data.

Several studies have improved structural decoders for long-sequence decoding and visual alignment.
Local attention~\cite{Nam23LA}, token compression~\cite{Zhu24}, and coverage modeling~\cite{COMER22} address the cost and instability of long sequences.
VAST~\cite{Huang23} aligns logical cell representations with image evidence using a visual alignment loss.

In contrast to prior work, this paper examines how the dependency topology of shared structural features affects downstream tasks.
Our approach leaves the output markup unchanged and does not rely on external OCR.
Instead, it refines structural representations obtained from the structural decoder with non-causal self-attention and cross-attention with image features.
This captures spatial and logical relationships among cells more precisely, leading to improved cell content recognition.

\section{Methodology}

\begin{figure*}[tb]
\centering
\includegraphics[width=\textwidth]{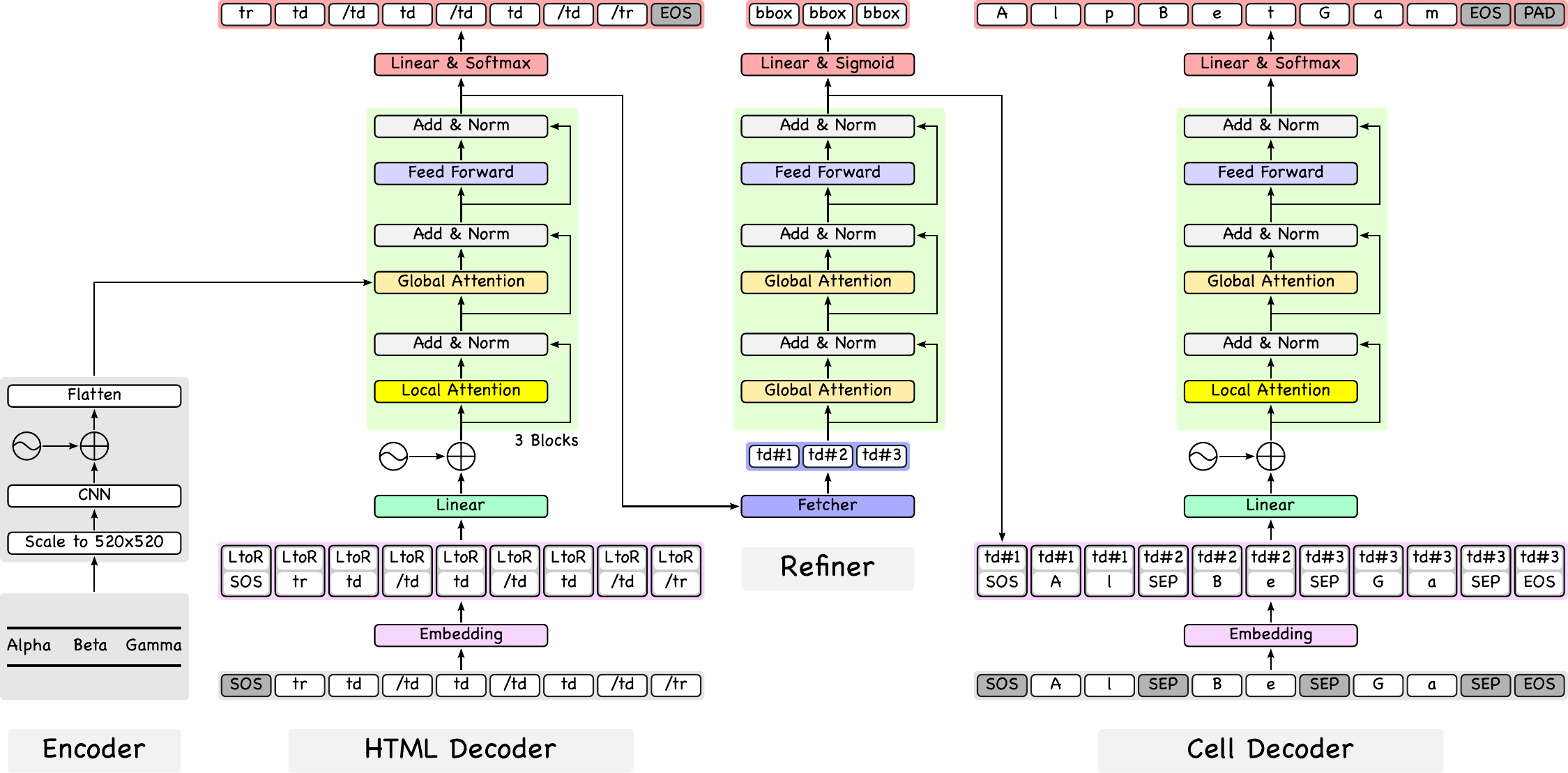}
\caption{Overview of the proposed multi-task framework.}
\label{fig:arch}
\end{figure*}

We adopt a multi-task model for joint structure prediction, cell localization, and cell content recognition.
Although prior work~\cite{Kat24,Nam23LA} has explored such multi-task models, it remains unclear what type of structural information should be shared across tasks.

Bounding boxes provide geometric supervision for cell localization but encode only position and size.
They do not directly encode inter-cell relationships, such as row-wise alignment, column-wise consistency, and global layout regularity.
In contrast, the hidden states of an HTML decoder implicitly accumulate structural information during decoding.
We therefore focus on extracting and refining such structural representations so that they can be effectively reused for downstream tasks.

However, causal sequence generation prevents earlier cell features from using information from later cells, limiting their structural context.
We thus decouple structural generation and structural aggregation.
The HTML decoder generates the structural skeleton, while a cell-level refiner reconstructs globally consistent representations through non-causal attention.

\subsection{Overall Architecture}

\figref{arch} provides an overview of the proposed architecture.
The model consists of a vision encoder, an HTML decoder, a cell-level refiner, a localization head, and a cell content decoder.
The encoder first extracts image features, and the HTML decoder generates HTML tokens sequentially conditioned on them.

Cell-level features associated with non-empty \TD{} tokens are extracted from the outputs of the HTML decoder.
The refiner then updates these features using non-causal self-attention and visual realignment.
The refined features are shared with both the localization head and the cell content decoder, enabling structural information to be aggregated independently of token generation.

\subsection{Cell-Level Structural Representation}

The decoder state corresponding to each non-empty \TD{} token is extracted and treated as the initial structural representation $\mathbf{h}_c$ of cell $c$.
These representations capture structural information accumulated during HTML decoding but remain conditioned on generation order.

Although AR decoding is required for sequential HTML generation, its causal constraint yields asymmetric structural representations when they are reused for downstream tasks.
This asymmetry may limit their suitability for cell localization and cell content recognition, where row and column dependencies are inherently bidirectional.

\subsection{Global Structural Refinement}

We introduce a dedicated refinement module that operates independently of the HTML decoder to construct order-independent and globally consistent structural features.
The refiner is implemented as a cell-level Transformer block consisting of self-attention, cross-attention, and a feed-forward network.

First, all cell representations are globally contextualized through non-causal multi-head self-attention:
\begin{equation}
\label{eq:MHSA}
\hat{\mathbf{h}}_c = \mathrm{MHSA}(\{\mathbf{h}_1,\dots,\mathbf{h}_C\})_c.
\end{equation}
In contrast to the HTML decoder, \eqref{MHSA} replaces the causal dependency pattern with fully connected cell-level self-attention, which enables the model to capture table-wide alignment and layout consistency.

Second, the resulting features are realigned with the full encoder feature map through multi-head cross-attention:
\begin{equation}
\label{eq:MHCA}
\bar{\mathbf{h}}_c = \mathrm{MHCA}(\hat{\mathbf{h}}_c, \mathbf{F}),
\end{equation}
where $\mathbf{F}$ denotes the image features from the encoder.
\eqref{MHCA} links the globally aggregated structural representations back to the image features, grounding the refined cell features in visual information related to text layout, ruling lines, and cell spacing.

Finally, a position-wise feed-forward network produces the refined structural feature:
\begin{equation}
\tilde{\mathbf{h}}_c = \mathrm{FFN}(\bar{\mathbf{h}}_c).
\end{equation}
The resulting representation $\tilde{\mathbf{h}}_c$ integrates inter-cell structural dependencies and image-grounded geometric information.
These refined representations are shared with both the localization head and the cell content decoder.

This refinement benefits both cell localization and content recognition.
When borders are invisible or cell contents span multiple lines, bounding box regression can be ambiguous.
The refined cell-level representations provide image-grounded structural context, promoting geometrically coherent localization and improving IoU.

The refiner module is designed to change the dependency structure of shared representations, rather than merely to add model capacity.
AR decoding restricts cell-level dependencies to a causal lower-triangular graph, whereas the proposed non-causal attention enables fully connected inter-cell dependencies.
We evaluate this topology-focused design choice empirically in \secref{ablation}.

\subsection{Parallel Cell Decoding}

\begin{algorithm}[tb]
\caption{Parallel cell decoding}
\label{alg:parallel}
\begin{algorithmic}[1]
\State Initialize $S$ with \texttt{SOS} and $C$ \texttt{SEP} tokens
\State Set all cells active
\For{$t=0,1,2,\ldots$}
  \State Predict next-token probabilities for each cell
  \For{each active cell $c$}
    \State Insert predicted token before cell $c$'s \texttt{SEP}
    \If{EOS emitted}
      deactivate cell $c$
    \EndIf
  \EndFor
  \If{all cells inactive}
    break
  \EndIf
\EndFor
\end{algorithmic}
\end{algorithm}

MuTabNet~\cite{Kat24} decodes cell contents sequentially across cells, allowing preceding cell contents to provide context for subsequent cells.
This design improves content recognition through cross-cell information sharing, but makes decoding cost scale linearly with the total number of cell-content tokens.
In our approach, structural dependencies are captured during the global refinement stage, reducing the need for sequential cross-cell conditioning during content generation.

We therefore decode cell contents independently and in parallel.
At each time step, all active cells predict their next tokens simultaneously.
Cells that emit an end-of-sequence token are deactivated, while the others continue decoding.
This procedure is summarized in \algoref{parallel}.
No cross-cell attention is performed at this stage, since the refined cell features already encode inter-cell relationships.

Under this design, cell decoding is bounded by the length of the longest cell sequence, rather than by the total number of cell-content tokens.
Since cells are decoded in parallel, short cells finish early and no longer contribute to subsequent iterations.
The refinement stage adds only a single attention block over cell-level features, with modest overhead compared with sequential concatenated content decoding.
This improves scalability while preserving recognition accuracy.

\section{Experiments}

We conducted experiments on two public table datasets.

\subsection{Datasets}

We utilized FinTabNet (FTN)~\cite{Zheng21} and PubTabNet (PTN)~\cite{Zhong20} for training and evaluation.
Both datasets support evaluation of table structure recognition, cell localization, and cell content recognition, and are regarded as de facto standard benchmarks for image-based table recognition.
Statistics are shown in \tabref{sets}.

\begin{table}[tb]
\centering
\caption{The statistics of the table image datasets.\label{table:sets}}
\medskip
\begin{tblr}{
colspec={lrrrrrr},
cell{1}{2,5}={c=3}{c}} \toprule
& Cells per table & & & Characters per cell \\ \cmidrule[r]{2-4} \cmidrule[l]{5-7}
Set    &  Train &  Valid &  Test & Train & Valid &  Test \\ \midrule
FTN    &  45.78 &  40.38 & 40.02 & 16.61 & 18.15 & 15.23 \\
PTN    &  64.86 &  66.56 & 72.52 & 13.37 & 13.54 & 12.36 \\
PTN250 & 144.45 & 146.42 &     - &  8.59 &  8.52 &     - \\ \bottomrule
\end{tblr}
\end{table}

\subsubsection{FinTabNet}

FTN is a large-scale dataset of table images from the annual reports of S\&P 500 companies.
It provides HTML tokens, cell bounding boxes, and cell contents as annotations.
The dataset contains approximately 112,000 tables and is split into training, validation, and test sets.
We used the \textit{validation} set, which contains 10,656 tables, for evaluation, following prior work~\cite{Zheng21,Nassar22,Nam23WS,Nam23GA,Nam23LA,Kat24}.
We also prepared four subsets by requiring at least 5, 10, 15, and 20 tokens per cell for the ablation study.

\subsubsection{PubTabNet}

PTN is a large-scale dataset of scientific tables from the PubMed Central Open Access subset.
It provides HTML tokens, cell bounding boxes, and cell contents as annotations.
The dataset contains approximately 510,000 images and is split into training and validation sets, but does not include an official test set.
The official test set was released by ICDAR for the competition~\cite{ICDAR21}.

PTN250~\cite{Nam23LA} is an unofficial subset of tables containing at least 250 structural tokens.
It is comparable in size to FTN and contains many cells per table, making it suitable for the ablation study.
Since only 76 samples had more than 15 content tokens per cell, we excluded subsets with higher token thresholds.

\subsection{Metrics}

Since a typical table has a nested structure of rows, cells, cell contents, and cell attributes, we employed TEDS~\cite{Zhong20} for evaluation.
We parsed the predicted and ground-truth HTML sequences into trees and calculated TEDS as follows:
\begin{equation}
\mathrm{TEDS}(T_1, T_2) = 1 - \frac{\TED{}(T_1, T_2)}{\max(|T_1|, |T_2|)},
\end{equation}
where $T_1$ and $T_2$ are the trees, $\TED$ is the tree edit distance function that reflects insertions, deletions, and renames, and $|T|$ is the number of nodes in $T$.

In addition, we report S-TEDS, which evaluates the predicted table structure without cell contents, whereas TEDS evaluates the complete table representation including cell contents.

Following the standard evaluation protocol~\cite{Zhong20}, tables are divided into simple and complex subsets.
Simple tables contain no vertically or horizontally merged cells, whereas complex tables contain at least one merged cell.

\subsection{Implementation}

We built our implementation on MMOCR~\cite{MMOCR} by extending MuTabNet~\cite{Kat24}.
We refer to this extended model as MuTabNet M2.

All models were trained for 30 epochs on four GPUs with a global batch size of 8 using the Ranger~\cite{RANGER19} optimizer.
The learning rate started at $10^{-3}$ and was reduced to $10^{-4}$ for epochs 26--28 and $10^{-5}$ for epochs 29--30.
Except for the full PTN setting, each experimental condition was trained three times with different random seeds.

Each image was padded with black pixels to preserve its aspect ratio, resized to $520\times{}520$ pixels, and normalized in RGB space.
Each cell bounding box was represented by its center coordinates and size, with all values normalized to the range $[0,1]$.
No data augmentation was applied for fair comparison with previous work.

Table images were encoded using the same backbone as MuTabNet~\cite{Kat24}.
Each token was converted into a 512-dimensional embedding.
The attention blocks in the two decoders and the refiner were configured with 8 heads.
The window size for local attention was set to 300.
We used greedy search with maximum lengths of 800 and 8000 for the HTML and cell decoders, respectively.

\subsection{Performance}

\begin{table}[tb]
\centering
\caption{Comparison on FinTabNet evaluation set.\label{table:fin}}
\medskip
\begin{tblr}{
colspec={lcccccc},
cell{1}{2}={c=3}{c},
cell{1}{6}={c=2}{c},
cell{3}{1}={r=2}{l},
cell{8}{1}={r=3}{l},
cell{12}{1}={r=2}{l},
cell{13}{5,7}={font=\bf}} \toprule
& Ablation & & & S-TEDS & TEDS \\ \cmidrule[r]{2-4} \cmidrule[lr]{5-5} \cmidrule[l]{6-7}
Model                & WS  & LA  & PT  & Total  & OCR & Total \\ \midrule
GTE$\dagger$         & -   & -   & -   & 87.14  & -   & -     \\
                     & -   & -   & \CM & 91.02  & -   & -     \\
TableFormer$\dagger$ & -   & -   & -   & 96.80  & -   & -     \\
VAST$\dagger$        & -   & -   & -   & 98.63  & \CM & 98.21 \\
Zhu et al.$\dagger$  & -   & -   & -   & 99.04  & \CM & 96.82 \\ \midrule
MTL-TabNet$\dagger$  & \CM & -   & -   & 98.72  & -   & 95.32 \\
                     & -   & -   & -   & 98.79  & -   & -     \\
                     & -   & \CM & -   & 98.85  & -   & 95.74 \\ \midrule
MuTabNet$\dagger$    & -   & \CM & -   & 98.87  & -   & 97.69 \\ \midrule
MuTabNet M2          & -   & \CM & -   & 98.87  & -   & 97.93 \\
                     & -   & \CM & \CM & 99.07  & -   & 98.08 \\ \bottomrule
\end{tblr}

\medskip
\begin{minipage}{.85\linewidth}
\footnotesize
$\dagger$ denotes results reported in the original papers.
WS and LA denote weak supervision and local attention.
PT denotes PubTabNet pretraining.
OCR denotes the use of an external OCR engine.
\end{minipage}
\end{table}

\begin{table}[tb]
\centering
\caption{Comparison on PubTabNet validation set.\label{table:val}}
\medskip
\begin{tblr}{
colspec={lccccccc},
cell{1}{2}={c=2}{c},
cell{1}{5}={c=4}{c},
cell{9}{4}={font=\bf},
cell{10}{1}={r=3}{l},
cell{14}{6-8}={font=\bf}} \toprule
& Ablation & & S-TEDS & TEDS \\ \cmidrule[r]{2-3} \cmidrule[lr]{4-4} \cmidrule[l]{5-8}
Model                & WS  & LA  & Total & OCR & Simple & Complex & Total \\ \midrule
EDD$\dagger$         & -   & -   & 89.90 & -   & 91.20  & 85.40   & 88.30 \\
TableFormer$\dagger$ & -   & -   & 96.75 & \CM & 95.40  & 90.10   & 93.60 \\
SEM$\dagger$         & -   & -   &     - & \CM & 94.80  & 92.50   & 93.70 \\
LGPMA$\dagger$       & -   & -   & 96.70 & \CM &     -  &     -   & 94.60 \\
VCGroup$\dagger$     & -   & -   &     - & \CM &     -  &     -   & 96.26 \\
VAST$\dagger$        & -   & -   & 97.23 & \CM &     -  &     -   & 96.31 \\
Zhu et al.$\dagger$  & -   & -   & 98.82 & \CM &     -  &     -   & 96.77 \\ \midrule
MTL-TabNet$\dagger$  & \CM & -   &     - & -   & 97.89  & 95.02   & 96.48 \\
                     & -   & -   &     - & -   & 97.92  & 95.36   & 96.67 \\
                     & -   & \CM &     - & -   & 98.07  & 95.42   & 96.77 \\ \midrule
MuTabNet$\dagger$    & -   & \CM & 97.71 & -   & 98.16  & 95.53   & 96.87 \\ \midrule
MuTabNet M2          & -   & \CM & 97.74 & -   & 98.34  & 95.68   & 97.04 \\ \bottomrule
\end{tblr}

\medskip
\begin{minipage}{.85\linewidth}
\footnotesize
$\dagger$ denotes results reported in the original papers.
WS and LA denote weak supervision and local attention.
OCR denotes the use of an external OCR engine.
\end{minipage}
\end{table}

\begin{table}[tb]
\centering
\caption{Comparison on ICDAR evaluation set.\label{table:test}}
\medskip
\begin{tblr}{
colspec={lcccccc},
cell{1}{2}={c=3}{c},
cell{1}{5}={c=3}{c},
cell{6}{1}={r=3}{l},
cell{10}{5-7}={font=\bf}} \toprule
& Ablation & & & TEDS \\ \cmidrule[r]{2-4} \cmidrule[l]{5-7}
Model                  & WS  & LA  & OCR & Simple & Complex & Total \\ \midrule
XM$\dagger$            & -   & -   & \CM & 97.60  & 94.89   & 96.27 \\
VCGroup$\dagger$       & -   & -   & \CM & 97.90  & 94.68   & 96.32 \\
Davar-Lab-OCR$\dagger$ & -   & -   & \CM & 97.88  & 94.78   & 96.36 \\ \midrule
MTL-TabNet$\dagger$    & \CM & -   & -   & 97.51  & 94.37   & 95.97 \\
                       & -   & -   & -   & 97.60  & 94.68   & 96.17 \\
                       & -   & \CM & -   & 97.77  & 94.58   & 96.21 \\ \midrule
MuTabNet               & -   & \CM & -   & 97.94  & 94.85   & 96.42 \\ \midrule
MuTabNet M2            & -   & \CM & -   & 98.10  & 95.08   & 96.62 \\ \bottomrule
\end{tblr}

\medskip
\begin{minipage}{.85\linewidth}
\footnotesize
$\dagger$ denotes results reported in the original papers.
WS and LA denote weak supervision and local attention.
OCR denotes the use of an external OCR engine.
\end{minipage}
\end{table}

We evaluate the proposed method against representative prior approaches under standard table recognition protocols.
Recent VLMs are excluded from the main comparison because they typically require substantially larger training data and broader task settings, as discussed in \secref{related}.

Unless otherwise stated, all models were trained without additional training data.
The proposed method does not rely on any external OCR for cell content recognition.
Results of the proposed method are averaged over three independent training runs, except for the full PTN setting.

\subsubsection{FinTabNet}

The proposed model achieves the highest TEDS among end-to-end models as reported in \tabref{fin}.
VAST~\cite{Huang23} reports slightly higher TEDS using an external OCR engine.
In contrast, the proposed model jointly performs structure prediction, cell localization, and cell content recognition within a single model.

Compared with previous multi-task approaches~\cite{Kat24,Nam23LA,Nam23GA,Nam23WS}, the refiner yields consistent gains in TEDS, suggesting that the improvement mainly comes from better cell content recognition.
Although the absolute TEDS gains are moderate due to metric saturation, the improvements are consistent across settings.

\subsubsection{PubTabNet}

The proposed model achieves competitive or superior performance across both simple and complex tables as shown in \tabref{val}.
It also outperforms prior multi-task methods~\cite{Kat24,Nam23LA,Nam23GA,Nam23WS}, particularly on complex tables, suggesting improved modeling of inter-cell dependencies.

\tabref{test} further compares our method with top-performing systems from the ICDAR competition~\cite{ICDAR21}.
The consistent improvements on both sets demonstrate the robustness of the refinement and parallel decoding strategy.
The MuTabNet results are recomputed after removing redundant \texttt{html} and \texttt{body} elements from the original TEDS evaluation~\cite{Kat24}.

\subsection{Inference Time}

\begin{table}[tb]
\centering
\caption{Inference time on NVIDIA V100 GPUs.\label{table:time}}
\medskip
\begin{tblr}{
colspec={cccccc},
cell{1}{3}={c=3}{c},
cell{3,9}{1}={r=6}{l},
cell{3,6,9,12}{2}={r=3}{c},
cell{8,14}{6}={font=\bf}} \toprule
& & Tasks \\ \cmidrule{3-5}
Set & Parallel & HTML  & Bbox & Cell & Time (s) \\ \midrule
FTN & -        & \CM   & -    & -    & 0.674 \\
    &          & \CM   & \CM  & -    & 0.679 \\
    &          & \CM   & \CM  & \CM  & 3.707 \\ \midrule
    & \CM      & \CM   & -    & -    & 0.787 \\
    &          & \CM   & \CM  & -    & 0.792 \\
    &          & \CM   & \CM  & \CM  & 1.300 \\ \midrule
PTN & -        & \CM   & -    & -    & 1.049 \\
    &          & \CM   & \CM  & -    & 1.056 \\
    &          & \CM   & \CM  & \CM  & 4.634 \\ \midrule
    & \CM      & \CM   & -    & -    & 1.150 \\
    &          & \CM   & \CM  & -    & 1.156 \\
    &          & \CM   & \CM  & \CM  & 1.506 \\ \bottomrule
\end{tblr}
\end{table}

\tabref{time} reports inference time for cumulative task settings, from table structure prediction only to the full setting with bounding box regression and cell content generation.
Inference time was measured during AR decoding without KV-cache using four NVIDIA V100 GPUs with a total batch size of 8.

The concatenated AR decoding introduced in MuTabNet~\cite{Kat24} scales with the total number of tokens across all cells.
Parallel cell decoding instead synchronizes generation across cells, making the number of AR steps depend on the maximum cell length rather than the sum of all cell lengths.

The proposed strategy reduces overall inference time by approximately three times, while maintaining accuracy.
The additional cost of the refinement module is small compared with token-level AR decoding.

\subsection{Ablation Study\label{sec:ablation}}

\begin{table}[tb]
\centering
\caption{Ablation study by minimum content tokens per cell.\label{table:abl}}
\medskip
\subfloat[TEDS.]{
\begin{tblr}{
colspec={cccccccccc},
cell{1}{3}={c=8}{c},
cell{2}{3}={c=5}{c},
cell{2}{8}={c=3}{c},
cell{4,7}{1}={r=3}{c},
cell{9}{3,4,5,6,7,8,9,10}={font=\bf}} \toprule
& & TEDS \\ \cmidrule{3-10}
& & FTN & & & & & PTN250 \\ \cmidrule[r]{3-7} \cmidrule[l]{8-10}
Model   & Refiner & 0+   & 5+   & 10+  & 15+  & 20+  & 0+   & 5+   & 10+  \\ \midrule
Bbox    & -       & 97.1 & 97.1 & 96.5 & 94.0 & 91.3 & 95.6 & 95.3 & 94.5 \\
        & Causal  & 97.2 & 97.2 & 96.6 & 94.1 & 91.5 & 95.5 & 95.1 & 94.7 \\
        & Global  & 97.2 & 97.2 & 96.6 & 94.3 & 91.6 & 95.7 & 95.4 & 94.7 \\ \midrule
Full    & -       & 97.9 & 97.9 & 97.5 & 96.0 & 94.3 & 95.8 & 95.5 & 94.8 \\
        & Causal  & 97.8 & 97.8 & 97.5 & 96.0 & 94.2 & 95.7 & 95.4 & 94.8 \\
        & Global  & 97.9 & 97.9 & 97.7 & 96.3 & 94.7 & 95.9 & 95.7 & 95.2 \\ \midrule
\end{tblr}
} \\
\subfloat[IoU.]{
\begin{tblr}{
colspec={cccccccccc},
cell{1}{3}={c=8}{c},
cell{2}{3}={c=5}{c},
cell{2}{8}={c=3}{c},
cell{4,7}{1}={r=3}{c},
cell{9}{3-10}={font=\bf}} \toprule
& & IoU \\ \cmidrule{3-10}
& & FTN & & & & & PTN250 \\ \cmidrule[r]{3-7} \cmidrule[l]{8-10}
Model   & Refiner & 0+   & 5+   & 10+  & 15+  & 20+  & 0+   & 5+   & 10+  \\ \midrule
Bbox    & -       & 55.5 & 55.4 & 52.5 & 46.5 & 43.2 & 51.0 & 51.2 & 45.6 \\
        & Causal  & 64.4 & 64.4 & 62.9 & 57.8 & 54.3 & 51.9 & 52.0 & 46.2 \\
        & Global  & 62.6 & 62.6 & 60.5 & 54.3 & 50.2 & 51.7 & 51.8 & 45.9 \\ \midrule
Full    & -       & 81.7 & 81.8 & 81.4 & 79.4 & 77.6 & 82.9 & 83.4 & 83.1 \\
        & Causal  & 84.8 & 84.9 & 84.6 & 82.9 & 81.5 & 86.1 & 86.5 & 86.4 \\
        & Global  & 85.5 & 85.6 & 85.4 & 83.8 & 82.4 & 86.3 & 86.7 & 86.7 \\ \midrule
\end{tblr}
}
\end{table}

The ablation study is structured along two independent axes.
The first axis varies the shared information, where \emph{bbox} uses only cell coordinates and \emph{full} uses dense features from the HTML decoder.
The second axis compares no refiner, a causal refiner, and the proposed global refiner.

The causal refiner is introduced as a variant with the same architecture and parameter count as the global refiner, while differing only in the attention mask configuration.
This controlled comparison isolates whether the gains come from increased model capacity or from restructuring the inter-cell dependency graph.

All variants were evaluated on FTN and PTN250 using a unified experimental protocol.
FTN contains relatively longer text content per cell, whereas PTN250 contains a larger number of cells per table.
We additionally analyze subsets with increasing minimum token length to examine behavior under stronger structural and geometric ambiguity.
\tabref{abl} reports TEDS and IoU for these variants and subsets.

\subsubsection{Effect of Structural Feature Sharing}

Comparing the bbox and full models shows that sharing dense structural features substantially improves localization accuracy.
The improvement is substantial on both datasets, with IoU increasing from 55.5 to 81.7 on FTN and from 51.0 to 82.9 on PTN250.

This gain does not come from feeding decoded cell contents back into the cell localization head.
Cell localization is performed directly from structural features prior to content decoding.
Instead, sharing the same cell-level features with the cell decoder allows the content recognition loss to shape the shared representation during training.
This auxiliary supervision encourages the shared representation to encode cell-specific visual cues, thereby improving the feature quality available to the localization head.

The slight TEDS improvement indicates that dense structural feature sharing benefits not only localization but also cell content recognition.
The improvement tends to become larger for longer cell contents, where accurate alignment between cell representations and content regions is more important.

\subsubsection{Effect of Global Feature Refinement}

Applying feature refinement improves localization under both shared-information settings.
In the bbox setting, adding the global refiner raises IoU from 55.5 to 62.6 on FTN and from 51.0 to 51.7 on PTN250.
In the full setting, it improves IoU from 81.7 to 85.5 on FTN and from 82.9 to 86.3 on PTN250.

To assess the role of dependency topology, we compare the causal and global refiners under the full setting.
The causal refiner already improves IoU over the no-refiner case, showing the benefit of feature refinement itself.
The global refiner further improves IoU over the causal refiner, from 84.8 to 85.5 on FTN and from 86.1 to 86.3 on PTN250.
For TEDS, the causal refiner tends to slightly degrade performance, whereas the global refiner consistently recovers or improves it.

In the bbox setting, both refiners improve over the no-refiner baseline, while the causal refiner achieves higher IoU than the global refiner on FTN.
Since the cell decoder receives only bounding-box features in this setting, the refined dense features are used mainly for localization.
This observation is consistent with the view that global refinement is more effective when the refined structural features are shared with both localization and content recognition, as in the full setting.

Since the causal and global refiners have identical architecture and parameter count, this gap supports our hypothesis that replacing the lower-triangular causal dependency structure with a fully connected inter-cell graph improves structural representation quality.
The advantage is more evident on subsets with longer cell contents, where geometric ambiguity and cross-cell dependencies are stronger.

\subsection{Case Study}

\begin{figure*}[tb]
\centering
\subfloat[Accepting only bounding boxes.]{
\begin{minipage}{.48\textwidth}
\centering
\includegraphics[width=\textwidth]{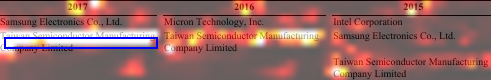} \\
\includegraphics[width=\textwidth]{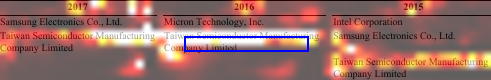} \\
\includegraphics[width=\textwidth]{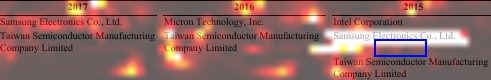} \\
\includegraphics[width=\textwidth]{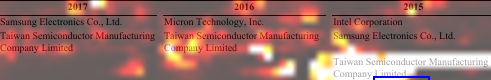}
\end{minipage}
}
\hfill
\subfloat[Accepting structural features.]{
\begin{minipage}{.48\textwidth}
\centering
\includegraphics[width=\textwidth]{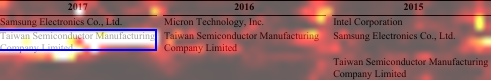} \\
\includegraphics[width=\textwidth]{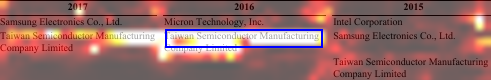} \\
\includegraphics[width=\textwidth]{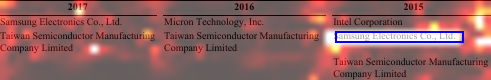} \\
\includegraphics[width=\textwidth]{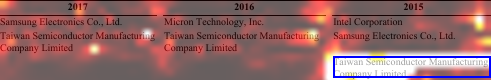}
\end{minipage}
}
\caption{Cell-decoder attention maps and predicted bounding boxes.\label{fig:case}}
\end{figure*}

\subsubsection{Contribution of Structural Features}

\figref{case} visualizes attention maps from the cell decoder and bounding boxes predicted from the refiner.
In the bbox-only setting, attention often spreads outside the target regions, leading to inaccurate bounding boxes and truncated multi-line recognition.
This suggests that sharing geometric coordinates alone is insufficient for coherent localization and decoding.

With shared and refined dense structural features, attention is more localized within the corresponding cell regions, and the predicted bounding boxes are more accurately aligned with cell boundaries.
The results in \secref{ablation} are consistent with these qualitative observations.

\begin{figure*}[tb]
\centering
\subfloat[Causal refiner.]{\includegraphics[width=.32\textwidth]{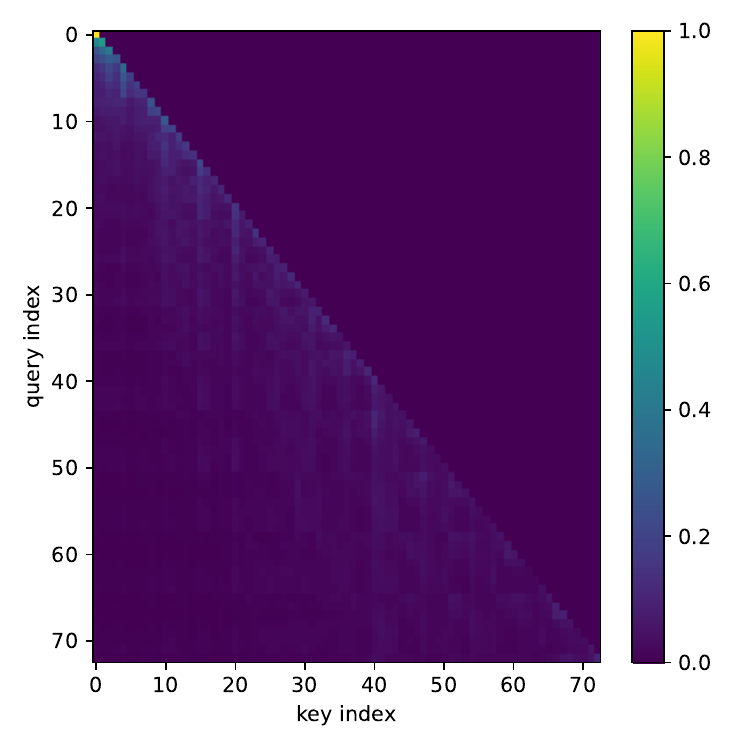}}
\subfloat[Global refiner.]{\includegraphics[width=.32\textwidth]{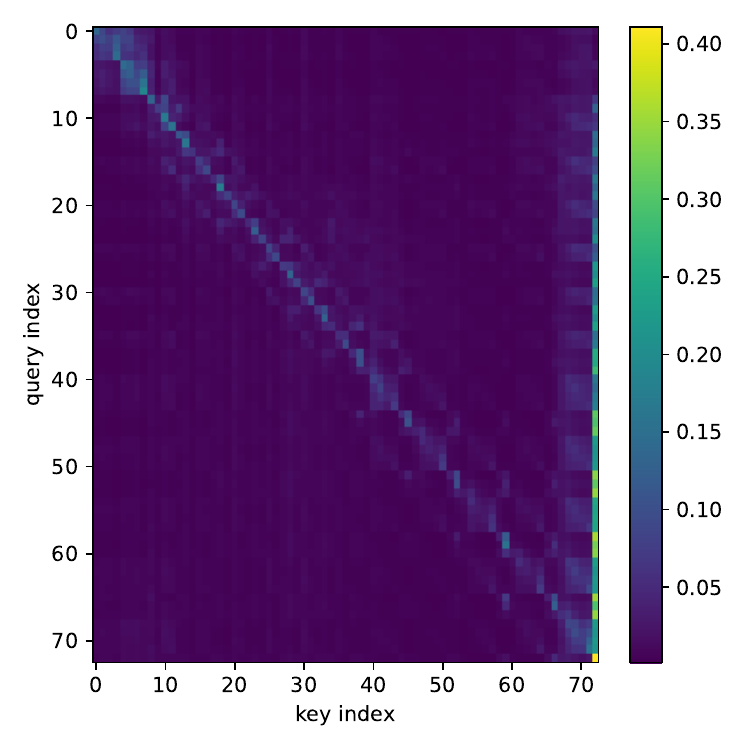}}
\subfloat[Entropy.]{\includegraphics[width=.32\textwidth]{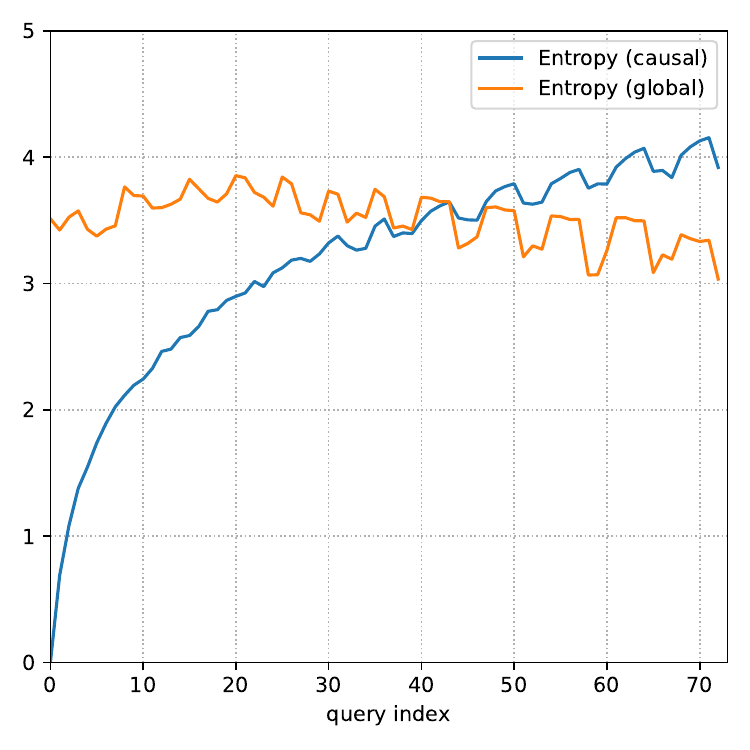}}
\caption{Self-attention maps and entropy profiles in the causal and global refiners.\label{fig:self}}
\end{figure*}

\subsubsection{Contribution of Global Refinement}

\figref{self} compares cell-level self-attention and entropy profiles in the causal and global refiners.
The causal refiner exhibits a lower-triangular dependency pattern due to AR masking, restricting each cell to attend only to previously generated cells.
The global refiner enables each cell to interact bidirectionally with all other cells, thereby forming a fully connected inter-cell dependency graph.

The entropy profiles further support this difference.
The causal refiner yields a position-dependent entropy profile, as later query positions have access to more keys.
In contrast, the global refiner exhibits a uniform entropy profile, indicating position-independent contextual access.

\subsection{Limitations}

The proposed refiner operates after the HTML decoder has generated the table structure.
It improves structural representations for cell localization and content recognition, but it does not directly revise structural errors such as missing cells or incorrect spans.
Such errors can degrade the final TEDS score.

\section{Conclusion}

We investigated how structural representations should be shared across tasks in table recognition.
We highlighted generation-order dependency in autoregressive decoder states as a limitation for cell localization and content recognition.

We introduced a structural refinement module that converts order-dependent decoder states into globally coherent cell-level structural representations.
Results on large-scale table datasets verify the effectiveness of the proposed method for both localization and end-to-end recognition.
The ablation study further suggests that modeling fully connected dependencies, rather than causal dependencies, is key to the observed improvements.
The proposed design also enables parallel cell decoding, reducing inference time by approximately threefold without degrading recognition accuracy.

These results underscore the role of dependency topology in shared structural representations and motivate explicit modeling of inter-cell dependencies.

{\small
\bibliographystyle{splncs04}
\bibliography{icdar26}
}

\end{document}